\documentclass[runningheads]{llncs}
\usepackage{graphicx}
\usepackage[tight,footnotesize]{subfigure} % for subfigures
\usepackage{epsfig}

\begin{document}
\title{Learning to Grasp from a Single Demonstration}
%
%\titlerunning{Abbreviated paper title}
% If the paper title is too long for the running head, you can set
% an abbreviated paper title here
%
\author{Pieter Van Molle\and
Tim Verbelen \and
Elias De Coninck \and \\
Cedric De Boom \and
Pieter Simoens \and
Bart Dhoedt}
\authorrunning{P. Van Molle et al.}
% First names are abbreviated in the running head.
% If there are more than two authors, 'et al.' is used.
%
\institute{Ghent University - imec, IDLab, \\
Department of Information Technology.\\
\email{firstname.lastname@ugent.be}}
\maketitle              % typeset the header of the contribution
\begin{abstract}
Learning-based approaches for robotic grasping using visual sensors typically require collecting a large size dataset, either manually labeled or by many trial and errors of a robotic manipulator in the real or simulated world. We propose a simpler learning-from-demonstration approach that is able to detect the object to grasp from merely a single demonstration using a convolutional neural network we call GraspNet. In order to increase robustness and decrease the training time even further, we leverage data from previous demonstrations to quickly fine-tune a GrapNet for each new demonstration. We present some preliminary results on a grasping experiment with the Franka Panda cobot for which we can train a GraspNet with only hundreds of train iterations.

\end{abstract}
\section{Introduction}

In the advent of Industry 4.0, more and more small and medium enterprises are looking into the adoption of robots to improve their production processes. Increasingly popular are the so-called collaborative robots or cobots. These robots are often lightweight and equipped with force torque sensors, enabling these robots to naturally stop in case of collisions, ensuring safety in human-robot collaboration scenarios \cite{Haddadin08}. Example cobots that are currently available off-the-shelf are the Kuka LBR series \cite{kuka}, the Universal Robots UR series \cite{ur} and the Franka Panda \cite{franka}, which can be used in a variety of applications such as production line loading and unloading, product assembly, and machine tending \cite{Bloss16}.

One way to program these collaborative robots is by kinesthetic demonstrations, in which the human operator takes the robot arm and moves it to the desired positions. This reduces the burden of programming the robot, as it is a much more intuitive approach and requires no expert knowledge on robot kinematics or programming code. Although in industry this approach is coined as ``learning from demonstration'', it is merely a record and replay feature as opposed to the learning from demonstration research in which generalized policies are trained using machine learning techniques \cite{Schaal96}. However, this limits the applicability of currently available systems to cases where positions are fixed relative to the robot. For example, when grasping an object, this object has to be at the same position for each repetition. Even a small perturbation of the object's position could harm the system.

In order to mitigate these limitations, one could attach a vision sensor to the robot and use this information to recognize objects, estimate their pose and calculate the best grasp \cite{Bohg2014}. However, these techniques require lots of supervision, grasp examples and/or training time. In this paper, we propose a grasping approach using neural networks that seamlessly fits in the current established workflow of programming a cobot, and requires only a single demonstration in order to allow perturbations in the grasping position of the target object, up to some extent.

The remainder of this paper is structured as follows. In the next section we give an overview of related work in robotic grasping and learning from demonstration. Next, we propose our approach, which we call GraspNet, in Section 3. We present some preliminary experimental results in Section 4. Finally we discuss our results and conclude with pointers for future work.

\section{Related work}

Grasping objects with a robotic manipulator is a long standing challenge in research~\cite{Nguyen86}. We focus on data-driven grasping, in which grasping is learned from vision data, either RGB images or depth scans. In particular, we distinguish three types of data-driven grasping: (i) using labeled training data, (ii) using human demonstrations and (iii) using trial and error. For a more in depth survey on data-driven grasping we refer to \cite{Bohg2014}.

The first type assumes a dataset is available with example objects and the corresponding grasp positions, either in the form of 3D meshes \cite{Goldfeder09} or 2D images \cite{Saxena08}. Next, a machine learning model such as a neural network can be trained to predict the correct grasp position \cite{Redmon14}. In practice this involves carefully collecting and labeling a dataset, on which training can be performed.

A second approach uses human demonstrations \cite{Tegin09}. In order to generalize well to situations that are different than the human demonstrations, this is often combined with reinforcement learning techniques \cite{Pastor09}, or guided policy search \cite{Levine16}. However, for this approach to work, quite a number of demonstrations is usually required in order to generalize well.

Reinforcement learning can also be used in the third type, in which grasping is learned purely from trial and error \cite{Peters08,Gu17}. Another trial and error approach was presented by Levine et al., in which 14 robots collect data over 800,000 grasp attempts \cite{Levine17}. Using this data, a convolutional neural network is trained to predict grasp success, and is then used to implement a controller. A similar approach was presented by Pinto et al. \cite{Pinto16}. These trials can also be executed in simulation, after which the learned policies are transferred to the real world \cite{Tobin17}. Although these approaches require the least amount of supervision, they are often impractical as they require a lot of trials before the first success.

In this paper we combine elements from the discussed approaches. Similar to \cite{Levine17}, we also train a neural network to predict grasp success, which we use to implement a grasp controller. However, instead of creating a large dataset using trial and error, we collect a single data point per demonstrated grasp. We show that it is possible to train a neural network that predicts the correct grasp position from this single data point through the use of various data augmentation techniques.

\section{Learning from a single demonstration}

\begin{figure*}[b!]
\centering
\subfigure[]{
\includegraphics[height=1.4in]{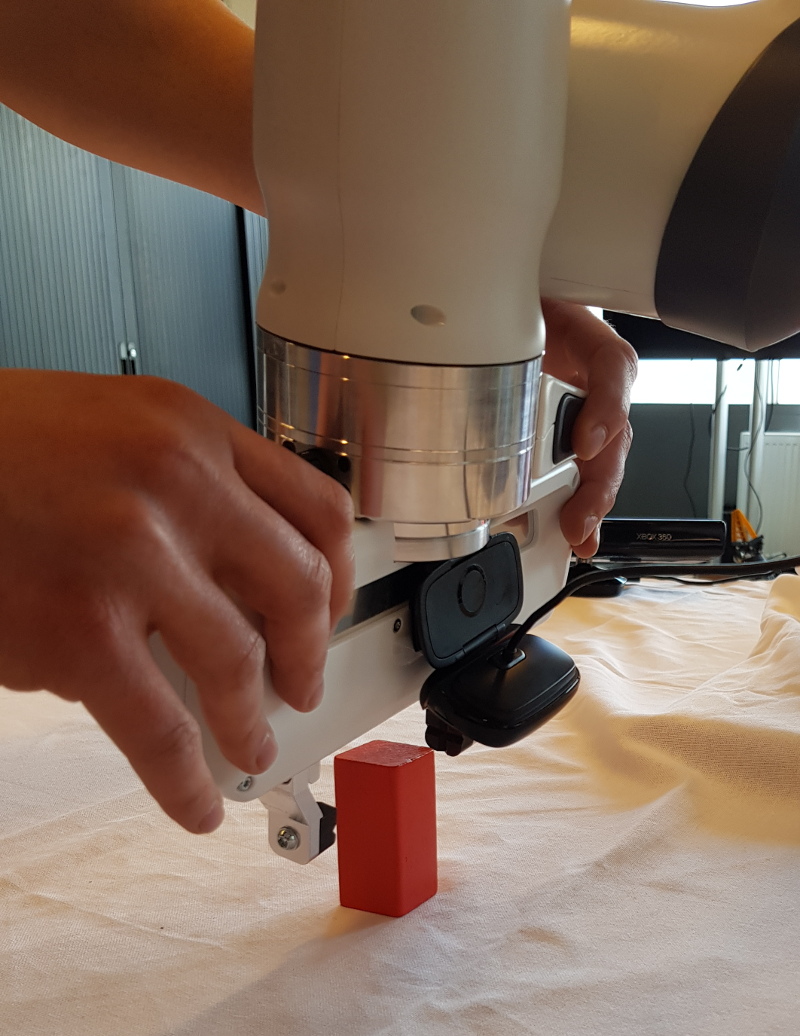}
\label{fig:teach}
}
\subfigure[]{
\includegraphics[height=1.4in]{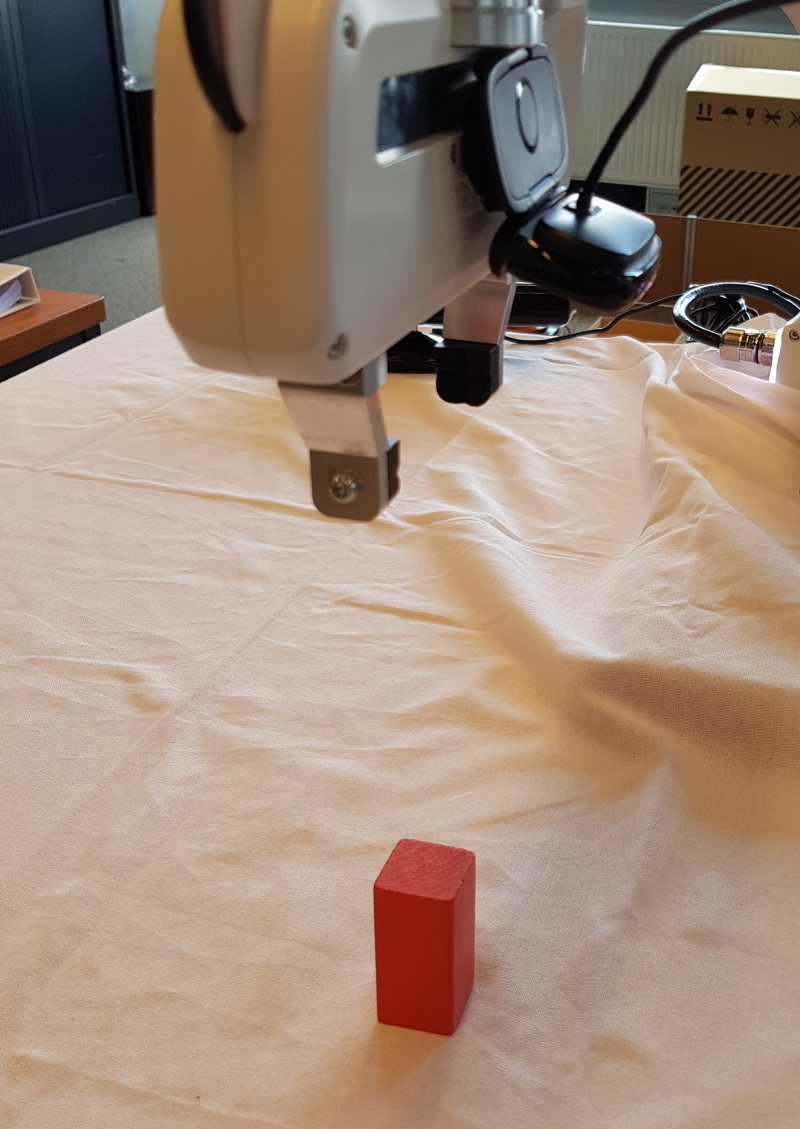}
\label{fig:record}
}
\subfigure[]{
\includegraphics[width=1in]{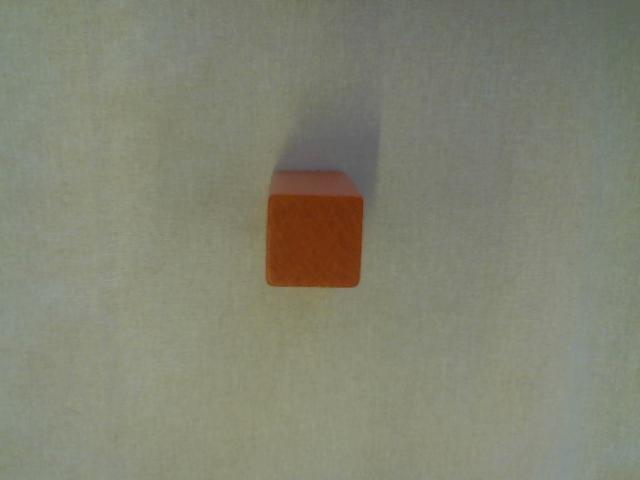}
\label{fig:base}
}
\subfigure[]{
\includegraphics[width=1in]{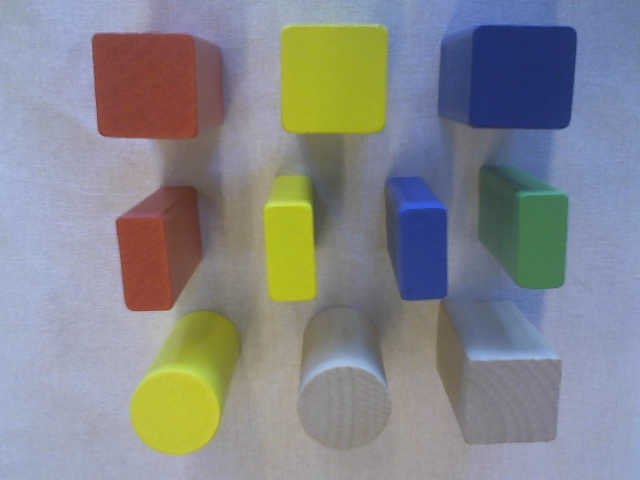}
\label{fig:all}
}
\caption{First, the operator guides the robot to the correct grasp pose (a). Next, the robot moves up (b) and the top-down camera captures a single frame used for training (c). We consider 10 types of blocks to grasp (d).}
\label{fig:setup}
\end{figure*}

Current collaborative manipulators have a so-called ``program by demonstration'' feature, with which the operator can easily program a sequence of actions for the cobot to execute, by guiding the end effector to the desired positions. This can, for example, be used to program simple pick-and-place tasks. However, at execution time, the robot will merely revisit the programmed positions, without any feedback or closed-loop control about whether the programmed task is actually succeeding. For example, when the object to grasp is not on the exact same position as during the demonstration, this will likely fail.

Our goal is to incorporate visual feedback, by mounting a camera on the end effector of the robot, training a neural network that identifies the object to grasp, and using a closed-loop control algorithm to execute the grasp. In order to mitigate the need of a large-scale dataset for training the neural network, we consider the following assumptions:
\begin{itemize}
	\item The robot operates in the same workspace as during the demonstrations.
    \item The object to grasp is the same as during the demonstration.
    \item The object to grasp can be grasped perpendicular to the workspace plane.
    \item The object to grasp has a positional offset of at most 8cm, with respect to the original position during demonstration.
\end{itemize}
Although these assumptions seem limiting, they still cover most use cases in an industrial environment, where objects need to be picked from and placed in well defined bin areas, but are not necessarily nicely aligned on fixed positions within these bins.

\subsection{Setup}

Our setup is shown in Figure~\ref{fig:setup} and consists of the Franka EMIKA Panda cobot, with a camera mounted on its end effector. To record a demonstration, the operator simply guides the gripper to the preferred grasp pose (a). Next, the robot will hover this position on a fixed height (b), and one camera frame is recorded (c). On this camera frame, the object to grasp will be positioned in the center of the image. As target objects, we currently use toy blocks of different shapes and colors. We gather demonstrations for the 10 types of blocks shown in (d).

\subsection{GraspNet}

\begin{figure}[t!]
\centering
\subfigure[]{
\includegraphics[width=1.4in]{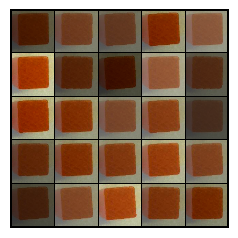}
\label{fig:positive}
}
\subfigure[]{
\includegraphics[width=1.4in]{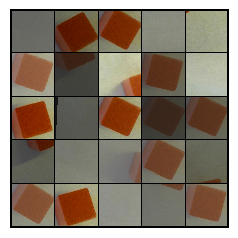}
\label{fig:negative}
}
\subfigure[]{
\includegraphics[width=1.4in]{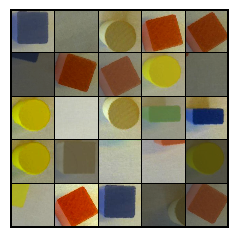}
\label{fig:negative_multi}
}
\caption{From the demonstration camera frame, we generate random positive (a) and negative (b) samples by cropping, rotating and adapting brightness and contrast. By including the demonstrations for the other blocks, we can generate additional negative samples (c).}
\label{fig:samples}
\end{figure}

\begin{figure}[b!]
\centering
\includegraphics[width=4.5in]{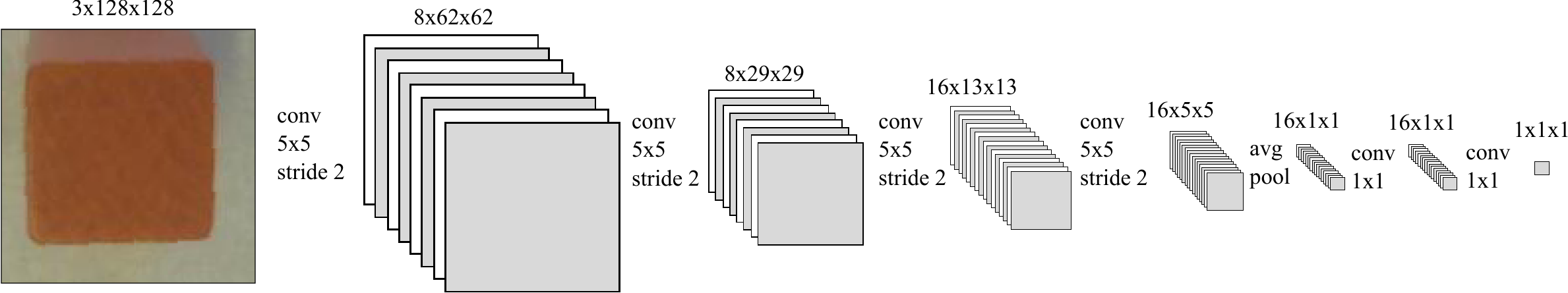}
\caption{GraspNet architecture: an $128 \times 128$ image crop is processed by 4 convolutional layers with 8, 8, 16 and 16 filters of size $5 \times 5$ and stride 2, followed by an average pooling layer and two fully connected layer with 16 hidden units. The fully connected layers are implemented as $1 \times 1$ convolutions and the output is a sigmoid neuron estimating grasp success.}
\label{fig:graspnet}
\end{figure}

We train a convolutional neural network, dubbed \textit{GraspNet}, to detect the correct grasp pose for a target object based on a camera frame, requiring but a single human demonstration. A well-known technique for extending a dataset and improving neural network training is data augmentation \cite{Chatfield14}, where images are perturbed to generate additional examples from the same underlying class. In this work, we rely on extreme data augmentation in order to create a ``very large'' dataset, starting from the single demonstration camera frame.

From the demonstration camera frame, with a resolution of $640 \times 480$ pixels, we generate a train set by taking random crops of $128 \times 128$ pixels. Positive samples are generated by taking center crops with a small random rotation (sampled uniformly between -3 and 3 degrees), while any other crop and rotation is used as negative sample. All samples are further randomized by applying random perturbations on brightness and contrast, both sampled uniformly between 0.5 and 1.5. Examples of positive and negative samples are shown in Figure~\ref{fig:samples}. 

Our neural network architecture is depicted in Figure~\ref{fig:graspnet}. It consists of four convolutional layers, of which the final layer is pooled using average pooling. We include two fully connected layers at the end of the network, implemented as $1 \times 1$ convolutions. All hidden layers have rectified linear units (ReLU), and the network ends with a single sigmoid neuron to classify each crop as positive or negative.

\subsection{Grasp controller}

\begin{figure}[b]
\centering
\includegraphics[width=\linewidth]{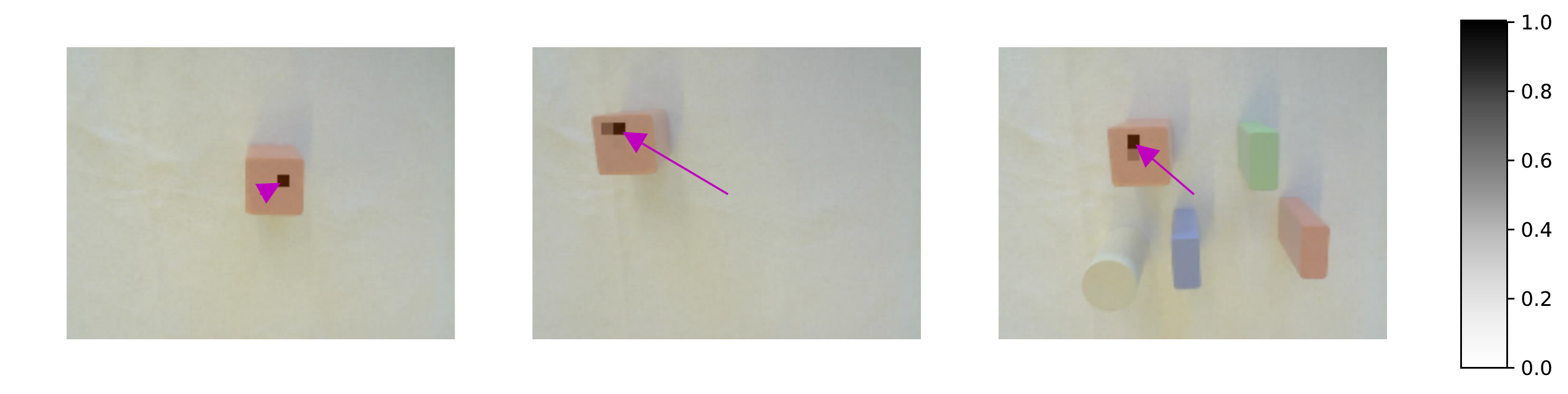}
\caption{Examples of camera images with activation map overlays, generated by running the images through GraspNet, and upscaling the output feature planes to match the input dimensions. Darker regions correspond to higher activations. The direction vector $v$ is represented by the magenta arrow. The big red block is always the target object.}
\label{fig:heatmap}
\end{figure}

By implementing the fully connected layers as $1 \times 1$ convolution kernels, we can easily apply this to an image with arbitrary size. The output is then a two-dimensional feature plane, which we interpret as an activation map that indicates the presence and position of the target grasp object, as seen in Figure~\ref{fig:heatmap}.

We implement a Cartesian velocity controller for the robot arm, that calculates the direction vector $v$ from the center of the image to the point of highest activation, and maps this to movement in the workspace plane. Once the highest activation is in the center of the image, the arm moves straight down and closes the gripper.

In order to also incorporate rotational information, we generate a batch of rotations of the camera image (e.g. ranging from -50 degrees to 50 degrees, with a step size of 10), and forward this batch through GraspNet, resulting in a batch of activation maps. Next, we rotate in the direction of the rotation with the highest maximum activation. This process is repeated until the unrotated image has the highest maximum activation.

By constantly streaming the camera images through GraspNet and acting accordingly we get a closed-loop controller that can successfully grasp the object once it is in the field of view of the camera.

\subsection{Grasping $n$ objects}

We can easily extend our approach when learning to grasp multiple object types, by providing a demonstration and training a separate GraspNet for each object type. However, when two objects look similar (for example the two red or blue blocks in our experiments), these might be difficult to distinguish, as we only train GraspNet from a single demonstration sample. In order to make GraspNet more robust, we combine all demonstration frames: when training for a certain target object, we extend our train set by regarding positive samples for other objects as negative samples for the target object. Examples of such negative samples are shown in Figure~\ref{fig:negative_multi}.

The main drawback of our approach is that we have to train a neural network from scratch for each demonstration. In order to mitigate this, we further improve our methodology by also learning a set of initialization parameters that can be fine-tuned quickly for a new demonstration. We apply the Reptile algorithm \cite{Nichol18}, a recently proposed approach for meta-learning for few-shot classification. Suppose we already have a set of $n$ demonstrations, we can train a set of initialization parameters $\phi$ by iteratively sampling one of the demonstrations, training GraspNet weights $W$ for this demonstration by applying $k$ gradient descent steps, and then updating $\phi \leftarrow \phi + \epsilon(W - \phi)$, where $\epsilon$ is the outer step size. This means we push $\phi$ to an optimum in which for all $n$ tasks we can get a high performing set of weights $W$ after $k$ additional gradient descent steps.

\section{Experiments}

\begin{figure}[t]
\centering
\includegraphics[width=\linewidth]{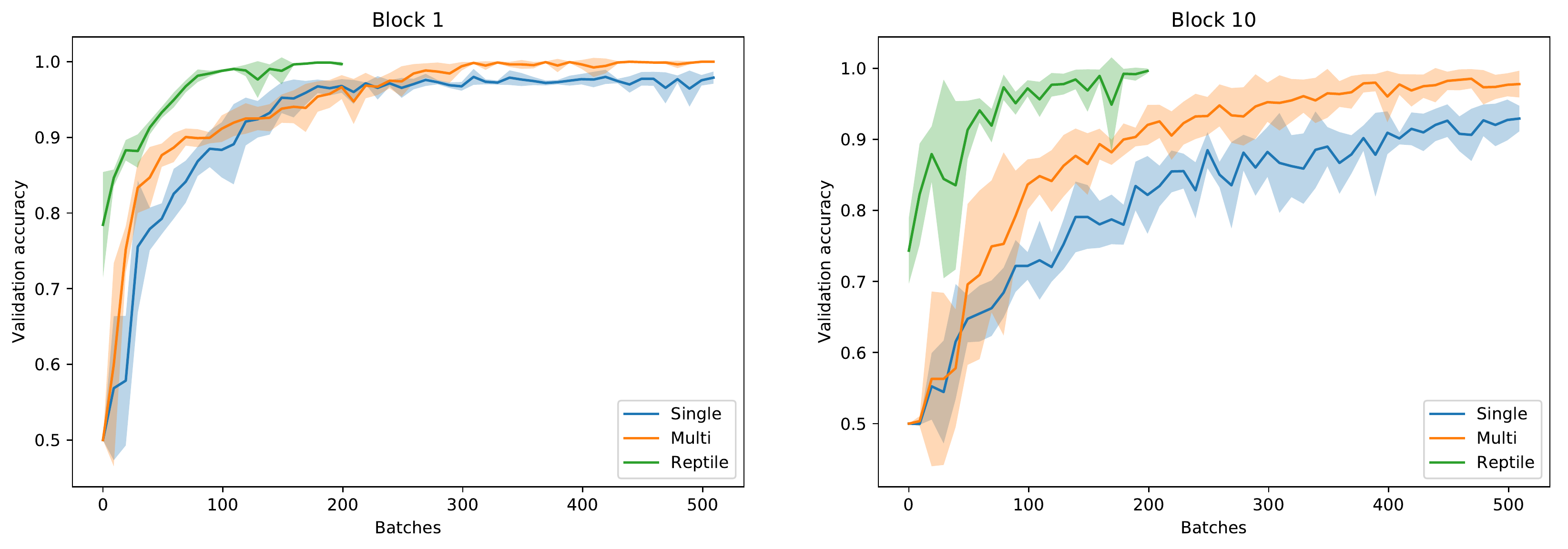}
\caption{Training performance for each configuration on an easy task (block 1) and a more difficult task (block 10). Averages over 5 random seeds are shown.}
\label{fig:training}
\end{figure}

We apply the dataset generation schemes described in Section~3 to generate a basic train set (using the single demonstration frame) and an extended train set (using all demonstration frames) for each of the 10 target blocks. Using these train sets, we train two GraspNet configurations for each block:  ``single'', in which the basic train set for the target block is used, and ``multi'', where the extended train set is used. Each GraspNet is trained for 512 iterations, with batches of 64 random samples. Parameter updates are done using the Adam algorithm \cite{Kingma2014}, with a step size of 0.001. We also implement the Reptile algorithm, for which we first train initialization parameters on nine of the demonstrations, and then evaluate for the remaining block. In this case, we first train a GraspNet for 250 outer iterations, with an initial outer step size of 0.6, which is linearly annealed to 0. We use meta-batches, where we train separately on four tasks, and average the update directions. Each task is trained for 10 iterations with a batch size of 10. Parameters are updated using the Adam algorithm, with a step size of 0.001, and $\beta_1 = 0$, to disable momentum, as done by \cite{Nichol18}.  Next, we train for 200 iterations of batch size 64 on the target task. All training is executed on an Nvidia Titan X GPU. The classification accuracy on a hold-out validation set (comprised of 200 random positive and 200 random negative samples, generated using the extended dataset generation scheme) for two blocks is plotted in Figure~\ref{fig:training}.

The results show that training using data of all demonstrations (multi) yields indeed better results, especially for the blocks that are more difficult to distinguish. We also see that training with Reptile indeed converges much faster, and in some cases even yields the best results.

We report the final classification accuracies for each of the configurations in Table~\ref{tab:results}. For each target block, we use a test set of 5,000 random positive and 5,000 random negative samples, again, generated using the extend dataset generation scheme. We see again that multi performs better or equal compared to single, but also that Reptile performs on par (or better), although this only requires a fraction of the training iterations for fine-tuning.

\begin{table}
\centering
% increase table row spacing, adjust to taste
\renewcommand{\arraystretch}{1.5}
\caption{Test set accuracies per configuration, for each target block, averaged over 5 random seeds.}
\label{tab:results}
\begin{tabular}{c c c c}
\hline
\textbf{Block} &\textbf{Single} &\textbf{Multi} &\textbf{Reptile} \\ 
\hline
1       &$\quad 0.989 \pm 0.004 \quad$    &$\quad \mathbf{0.999 \pm 0.001} \quad$    &$\quad 0.997 \pm 0.001 \quad$ \\
2       &$\quad 0.985 \pm 0.009 \quad$    &$\quad \mathbf{0.991 \pm 0.015} \quad$    &$\quad 0.989 \pm 0.007 \quad$ \\
3       &$\quad \mathbf{1.000 \pm 0.000} \quad$    &$\quad 0.995 \pm 0.010 \quad$    &$\quad 0.958 \pm 0.007 \quad$ \\
4       &$\quad 0.976 \pm 0.015 \quad$    &$\quad 0.993 \pm 0.009 \quad$    &$\quad \mathbf{0.999 \pm 0.000} \quad$ \\
5       &$\quad 0.969 \pm 0.021 \quad$    &$\quad \mathbf{0.999 \pm 0.001} \quad$    &$\quad 0.983 \pm 0.003 \quad$ \\
6       &$\quad 0.995 \pm 0.008 \quad$    &$\quad \mathbf{0.959 \pm 0.072} \quad$    &$\quad 0.954 \pm 0.046 \quad$ \\
7       &$\quad 0.910 \pm 0.055 \quad$    &$\quad \mathbf{0.994 \pm 0.007} \quad$    &$\quad 0.938 \pm 0.077 \quad$ \\
8       &$\quad 0.957 \pm 0.007 \quad$    &$\quad \mathbf{0.997 \pm 0.002} \quad$    &$\quad 0.974 \pm 0.024 \quad$ \\
9       &$\quad 0.887 \pm 0.035 \quad$    &$\quad \mathbf{0.960 \pm 0.021} \quad$    &$\quad 0.925 \pm 0.068 \quad$ \\
10      &$\quad 0.927 \pm 0.018 \quad$    &$\quad 0.974 \pm 0.022 \quad$    &$\quad \mathbf{0.993 \pm 0.002} \quad$ \\
\hline
\end{tabular}
\end{table}

\section{Discussion}

We also conduct some initial experiments using the real robot controlled by an Intel NUC equipped with an Intel i5-5250U CPU. When we do not take rotation into account, the robot is able to robustly pick up the block using GraspNet. When applying random rotations, we generate a batch of nine planes where we rotate the camera frame in steps of 10 degrees, and feed it through GraspNet. In this case, our robot successfully detects the correct grasp orientation, but the closed-loop controller fails to move the gripper to the exact grasp location. This is due to the fact that processing the batched input takes up to 700ms to process which is too high for accurate control. We are now looking into extending our setup using a separate, GPU-enabled machine for processing the camera frames.

Besides these promising results, we are aware of the current limitations of our approach. For example, as we only use a few data examples, this reduces the applicability to new and unseen situations. Of course this can be mitigated by adding additional data samples, or by using available grasp datasets \cite{Saxena08}. This would be especially appealing for usage of the algorithm, in which we can train once on a large dataset to obtain suitable initialization parameters to quickly fine-tune a GraspNet for each new demonstration.

One of the biggest advantages of our approach is that we can integrate it seamlessly with the currently common programming-by-demonstration workflow. A non-technical operator can still easily program the robot, while alleviating the need of supplying objects to grasp on the exact same position.

\section{Conclusion}

In this paper, we have proposed a technique for learning to grasp an object from a single demonstration from a vision sensor. For each demonstration, we trained a GraspNet, a convolutional neural network that predicts grasp success based on an image patch. Using such a GraspNet, we developed a closed-loop robot controller to grasp the object whenever it is in view of the camera. We show that we can in fact train a GraspNet from a single image, and that we can increase robustness by also including data from a few other demonstrations. Moreover, using the Reptile meta-learning algorithm, we show that we can quickly fine-tune a GraspNet for a new demonstration.

In future work, we plan to extend our approach to grasp more common objects instead of merely toy blocks, and integrate GraspNet in a programming-by-demonstration platform for a real collaborative industrial workspace. This will allow to test our approach on production-grade hardware platforms.

\section*{Acknowledgements}

Cedric De Boom is funded by a PhD grant of the Research Foundation - Flanders (FWO). We would like to thank Nvidia Corporation for their generous donation of a Titan X GPU.

%
% ---- Bibliography ----
%
% BibTeX users should specify bibliography style 'splncs04'.
% References will then be sorted and formatted in the correct style.
%
\bibliographystyle{splncs04}
\bibliography{bibtex}

\end{document}